\documentclass[lettersize,journal]{IEEEtran}
\usepackage{amsmath,amsfonts}
\usepackage{algorithmic}
\usepackage{algorithm}
\usepackage{array}
\usepackage[caption=false,font=normalsize,labelfont=sf,textfont=sf]{subfig}
\usepackage{textcomp}
\usepackage{stfloats}
\usepackage{url}
\usepackage{verbatim}
\usepackage{graphicx}
\usepackage{cite}

\usepackage{multirow}
\usepackage{bbding}
\usepackage{times}
\usepackage{epsfig}
\usepackage{amsmath}
\usepackage{amssymb}
\usepackage{caption}
\usepackage{booktabs} 
\usepackage{dsfont}
\usepackage{xcolor}
\usepackage[backref]{hyperref} 

\begin{document}

\title{P2PFormer: A Primitive-to-polygon Method for Regular Building Contour Extraction from Remote Sensing Images}

\author{Tao~Zhang, Shiqing~Wei, Yikang~Zhou, Muying~Luo, Wenling~Yu, and~Shunping~Ji
	
\IEEEcompsocitemizethanks{\IEEEcompsocthanksitem Tao Zhang, Yikang Zhou, Muying Luo and Shunping Ji are with the School of Remote Sensing and Information Engineering, Wuhan University, 129 Luoyu Road, Wuhan 430079, China.
\IEEEcompsocthanksitem Shiqing Wei is with the College of Oceanography and Space Informatics, China University of Petroleum (East China), Qingdao 266580, China.
\IEEEcompsocthanksitem Wenling Yu is with the School of Surveying and Geoinformation Engineering, East China University of Technology, Nanchang, 330013, China.
    \IEEEcompsocthanksitem  Corresponding author: Shunping Ji (jishunping@whu.edu.cn).}

}




\maketitle

\begin{abstract}
Extracting building contours from remote sensing imagery is a significant challenge due to buildings' complex and diverse shapes, occlusions, and noise. Existing methods often struggle with irregular contours, rounded corners, and redundancy points, necessitating extensive post-processing to produce regular polygonal building contours. To address these challenges, we introduce a novel, streamlined pipeline that generates regular building contours without post-processing. Our approach begins with the segmentation of generic geometric primitives (which can include vertices, lines, and corners), followed by the prediction of their sequence. This allows for the direct construction of regular building contours by sequentially connecting the segmented primitives. Building on this pipeline, we developed P2PFormer, which utilizes a transformer-based architecture to segment geometric primitives and predict their order. To enhance the segmentation of primitives, we introduce a unique representation called group queries. This representation comprises a set of queries and a singular query position, which improve the focus on multiple midpoints of primitives and their efficient linkage. Furthermore, we propose an innovative implicit update strategy for the query position embedding aimed at sharpening the focus of queries on the correct positions and, consequently, enhancing the quality of primitive segmentation. \textcolor{black}{Our experiments demonstrate that P2PFormer achieves new state-of-the-art performance on the WHU, CrowdAI, and WHU-Mix datasets, surpassing the previous SOTA PolyWorld by a margin of 2.7 AP and 6.5 AP$_{75}$ on the largest CrowdAI dataset. The code is available at \url{https://github.com/zhang-tao-whu/P2PFormer}.}
\end{abstract}

\begin{IEEEkeywords}
Regular building contour extraction, Primitive-based, Transformer, Remote sensing images
\end{IEEEkeywords}

\section{Introduction}
\label{sec:intro}
\begin{figure}[t]
  \centering
   \includegraphics[width=0.88\linewidth]{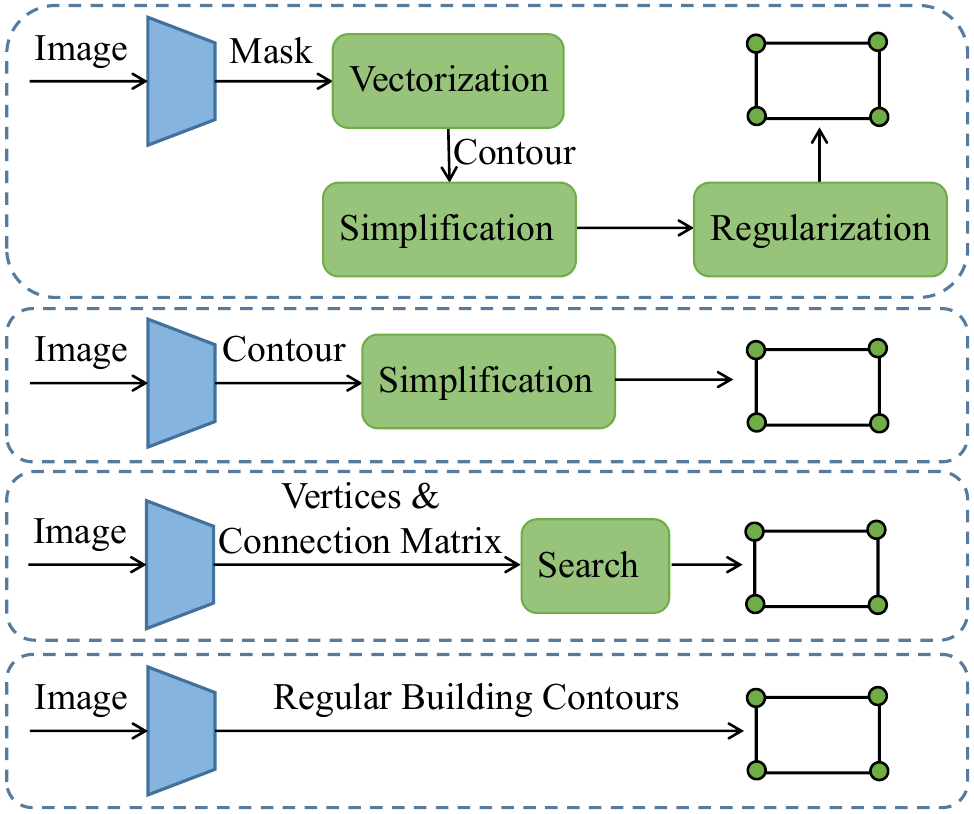}
   \caption{Different pipelines for regular building contour extraction. From top to bottom: mask-based, contour-based, vertex-based, and our proposed primitive-based pipeline. Blue trapezoids represent networks, and green rectangles represent post-processing steps. Only the end-to-end primitive-based pipeline completely gets rid of the handcrafted post-processing.}
   \label{fig:pipelines}
\end{figure}

The extraction of regular polygonal building contours from aerial or satellite imagery is vital for various applications, including cartography, urban planning, population estimation, and disaster management. However, current methods struggle to directly predict regular building contours, creating a discrepancy between the needs of downstream tasks and their capabilities.

There are three main-stream technical pipelines for extracting regular polygonal building contours: mask-based, contour-based, and vertex-based methods (a special case of primitive-based methods), as depicted in Figure \ref{fig:pipelines}. Mask-based methods \cite{mafcn, ffl} first segment buildings and then transform the segmentation results into regular building contours using vectorization and regularization algorithms. However, these methods often produce poor contour quality and internal holes in the segmentation result, leading to less-than-optimal performance. To address these issues, contour-based methods \cite{buildmapper, clpcnn} directly regress building contours, but they generate a large number of redundant points (\textit{e.g.}, 128) to model the contours of buildings with diverse shapes. Removing these redundant points using simplification algorithms is challenging, often resulting in imperfect results. On the other hand, vertex-based methods \cite{polyworld} avoid introducing redundant points by exclusively segmenting all building vertices in an image and regressing the connection matrix of the vertices. However, these methods still require post-processing algorithms as the connection matrix necessitates complex search algorithms to generate the corner order for each building. Furthermore, vertices can be easily missed due to their small size and potential for complete occlusion, which also hampers the performance of vertex-based methods.

In this paper, we propose a concise and elegant technical pipeline based on generic geometric primitives (as shown at the bottom of Figure \ref{fig:pipelines}), which can directly extract regular building contours by simultaneously predicting the positions and the order of primitives for each building. Following this primitive-based approach, we introduce P2PFormer (Primitive-to-Polygon using Transformer). P2PFormer comprises a detector for identifying the bounding boxes of buildings, an innovative primitive segmenter for segmenting primitives within a building's bounding box, and a novel primitive order decoder to determine the sequence of primitives. It is worth mentioning that P2PFormer can utilize any primitive, including not only vertex used in previous studies, but also straight line and corner, greatly expanding the scope of research.

The primitive segmenter, which follows a DETR-like architecture \cite{detr}, segments primitives based on the features of each building. In this segmenter, queries are used as primitive representations, allowing for the uniform formatting of arbitrary primitives. The segmenter utilizes cross-attention to enable queries to interact with the image and self-attention to model the relationships between primitives. To enhance the quality of primitive segmentation, we introduce three innovative designs: 1) We restrict the queries to interact solely with building features with a fixed sequence length in the bounding box (obtained through ROI-Align \cite{maskrcnn} from the image features) for each building. This resolves the issue of incorrectly connecting vertices from different buildings, as observed in \cite{polyworld}, and addresses the challenge of adjacent buildings with shared edges. 2) We introduce a novel representation for universal primitives called group queries. A primitive is represented by a group of queries and a single query position embedding, enabling a more effective feature representation of primitives that consist of multiple endpoints with possible long distances. 3) We dynamically update the query position embedding to expedite the focus of queries on the relevant position (e.g., midpoints of primitives), leading to greater accuracy in predicting the endpoint positions of primitives. These designs allow our approach to represent and model universal primitives effectively, thereby significantly improving the quality of primitive segmentation.

The primitive order decoder introduces a novel and straightforward approach that predicts the relative order of primitives directly, eliminating the need for any post-processing. To our surprise, we discovered that the primitive queries generated by the primitive segmenter are rich in information, already encompassing the positions and relationships of the primitives. Consequently, the primitive decoder employs only a few self-attention layers to reinforce the adjacency relationship between primitives. Following this, a single layer of Feed-Forward Network (FFN) is used to ascertain the sequence of primitives, yielding satisfactory results directly.

In summary, our main contributions are as follows:
\begin{itemize}
\item We introduce a novel primitive-based pipeline for directly extracting regular building contours that eliminates the need for post-processing. This method constructs contours from primitives using a bottom-up approach.
\item Following this pipeline, we have developed P2PFormer, which features a novel primitive segmenter and order decoder. The primitive segmenter includes three innovative designs that significantly enhance the quality of primitive segmentation.
\item \textcolor{black}{We evaluate vertex, line, corner as primitives for building contour extraction for the first time. We have conducted extensive experiments on the WHU, CrowdAI, and WHU-Mix datasets. The results show that P2PFormer achieves SOTA performance across all datasets.}
\end{itemize}

\section{Related Work}
\label{sec:rela}

\noindent\textbf{Mask-based Methods.} Prior studies \cite{r3, r4, r5, r6, r7, r8, r9} have concentrated on enhancing segmentation quality by using FCN \cite{fcn} as the foundational framework for building pixel classification. For instance, \cite{r3} employs a scale-robust CNN architecture to boost the precision of building segmentation, while \cite{r7} introduces a contour-guided and local structure-aware framework to improve the accuracy of building boundary extraction. However, these methods necessitate specific post-processing to extract the location information of building instances from the semantic segmentation map. Other studies \cite{mafcn, r16, r17, r18, polytransform, r20, ffl} generate initial contour polygons of building instances based on the connected foreground regions of the semantic segmentation map and refine these initial contours using various techniques such as handcrafted regularization algorithms, transformers, or frame field branches. However, the resulting building contours, often represented in masks, may exhibit zigzagged, irregular, and broken boundaries, which fall short of the manual delineation level required for downstream tasks~\cite{mafcn}.

\noindent\textbf{Contour-based Methods.} Contour-based methods aim to directly predict building contours, addressing the challenges that mask-based methods often encounter. Studies \cite{r23, r24, r25, r26, r27} utilize an RNN \cite{r28} to generate building contours by sequentially predicting vertices in a fixed direction. However, these methods frequently suffer from subpar prediction quality and accumulated errors. In contrast, \cite{clpcnn, r34} implement a two-stage pipeline for building contour extraction, which includes initial contour generation and subsequent contour refinement. Furthermore, \cite{buildmapper} employs a more robust contour extraction network supplemented with an additional point reduction head. While these contour-based methods effectively address zigzagged, irregular, and broken boundaries, they often necessitate specialized simplification post-processing to yield regular contours.

\noindent\textbf{Primitive-based Methods.} Prior research by \cite{polyworld} negates the necessity for simplification post-processing by employing a CNN to segment all building vertices and a Graph Neural Network (GNN) to forecast their interconnection relationships. However, they still depend on search algorithms to secure vertices and connection order for each building from an extensive connectivity matrix. Moreover, all modules in \cite{polyworld} are explicitly designed for the vertex primitive, which is not optimal due to its vulnerability to occlusion. In this paper, we introduce P2PFormer, a method that entirely eradicates the need for post-processing. It first detects building instances, then segments general primitives like vertices, lines, and corners, and directly regresses the primitives' order within the bounding box of each building.

\section{Method}
\label{sec:method}

\begin{figure*}[t]
\centering
\includegraphics[width=0.88\linewidth]{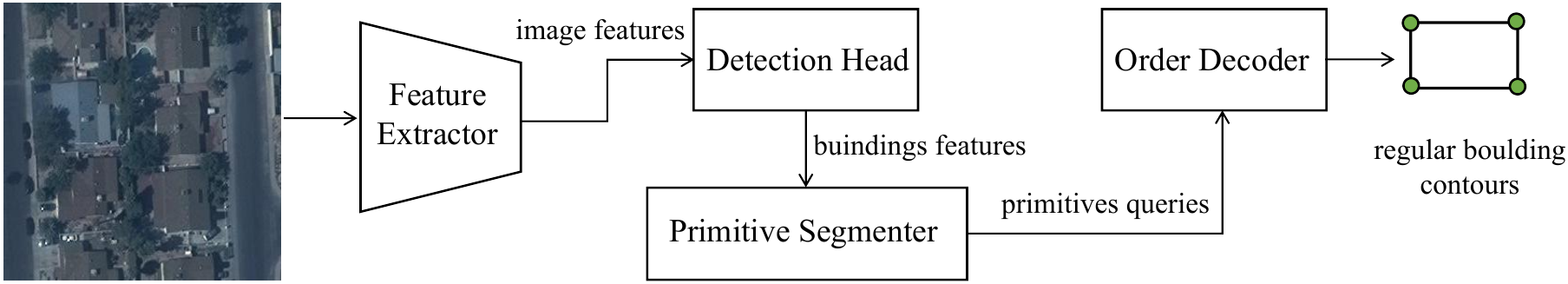}
\caption{Overall pipeline of P2PFormer. Initially, P2PFormer detects each building. Subsequently, it carries out primitive segmentation and primitive order regression, utilizing the building feature.}
\label{fig:overall pipeline}
\end{figure*}

P2PFormer introduces a streamlined and elegant architecture that hinges on the fundamental concept of directly generating regular building contours from primitives with topological orders, eliminating the need for any post-processing. The comprehensive workflow of P2PFormer is depicted in Figure \ref{fig:overall pipeline}. Initially, the bounding box for each building is produced using a conventional detection head; in this research, the FCOS head is employed as referenced in~\cite{fcos}. Subsequently, the primitive segmenter segments a predetermined number of primitives, predicting their positions and confidence scores within the bounding box of each building. Erroneous and superfluous primitives are efficiently filtered out based on the predicted confidence scores. Ultimately, the order decoder ascertains the sequence of the retained primitives, drawing on the queries of the primitives. The subsequent sections provide a detailed exposition of these processes.

\subsection{Primitive Segmenter}

\begin{figure*}[t]
\centering
\includegraphics[width=0.75\linewidth]{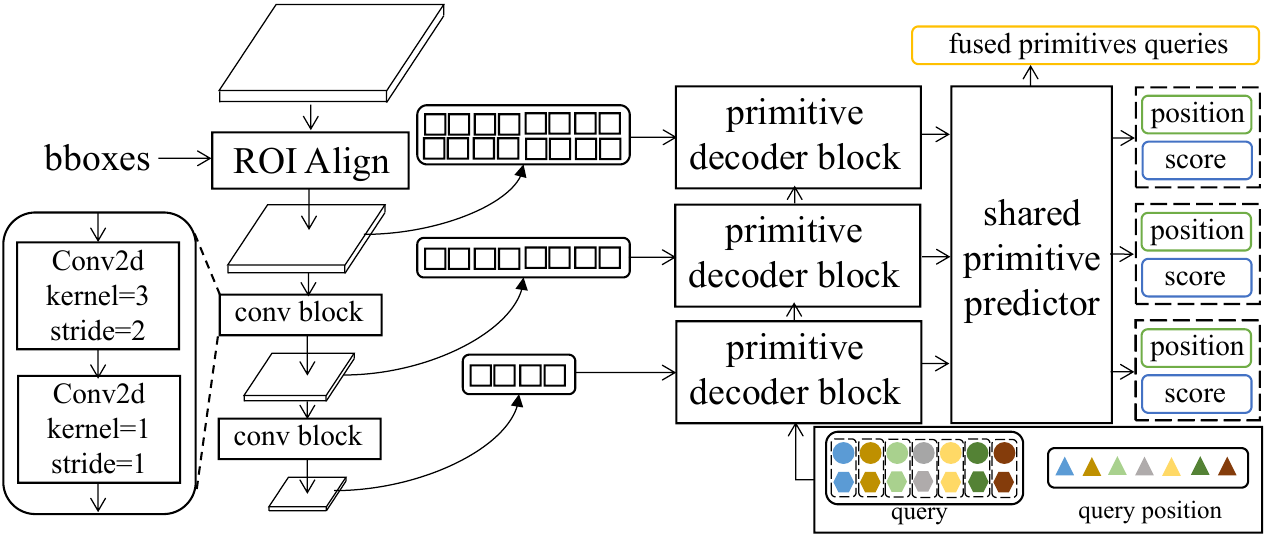}
\caption{Architecture of the primitive segmenter. Circles and hexagons represent group queries, while triangles denote query position embeddings. The same color signifies that they belong to the same primitive.}
\label{fig:primitive segmenter}
\end{figure*}

The architecture of the primitive segmenter, as depicted in Figure \ref{fig:primitive segmenter}, consists of stacked primitive decoder blocks. These blocks utilize standard cross-attention to gather information from multi-scale building features and standard self-attention to model relationships among primitives within a building. The positions and confidence scores of the primitives are predicted using a Multilayer Perceptron (MLP) and Feed-Forward Network (FFN) based on their representations.

\noindent\textbf{Multi-Scale Input Sequence Generation.} Fixed-size building features are generated from image features using ROI-Align~\cite{maskrcnn} to crop them according to bounding boxes. These features are subsequently transformed into multi-scale features via a series of 2$\times$ down-sampling blocks. After flattening, they serve as inputs for the respective decoder blocks, which helps to decrease memory consumption and accelerate network convergence. The initial decoder block utilizes the smallest-size feature, which contains denser semantic information, to rapidly direct the query to the pertinent area. Conversely, the final decoder block employs the largest-size feature, enriched with detailed information, to enhance the precision of the prediction.

\begin{equation}
\label{eq:1}
\left\{
\begin{aligned}
   &f^{instance}_{i+1}=DownSample(f^{instance}_{i}) \\
   &f^{decoder}_{i}=Flatten(f^{instance}_{N_{sc} + 1 - i})
\end{aligned}
\right.,
\end{equation}
\textcolor{black}{where $N_{sc}$ represents the number of scales (here 3), $f^{instance}_{i}\in \mathbb{R}^{\frac{S}{2^i} \times \frac{S}{2^i} \times C} (i\in{1,2,3})$ is the \emph{i}-th instance feature with C channels, and $f^{decoder}_{i}\in \mathbb{R}^{\frac{S^2}{4^i} \times C}$ is the input feature of \emph{i}-th primitive decoder block.}

\noindent\textbf{Primitive Representation.} A single query can only attend to a small portion of the image \cite{dabdetr}. However, when using lines or corners as primitives, multiple points must be attended to simultaneously, which cannot be achieved well with a single query. In contrast to previous methods such as \cite{detr, letr}, we propose a novel representation strategy where each primitive is represented by a group of queries and a single query position embedding. Specifically, $query_{i} \in \mathbb{R}^{n\times C} $ in Equation \ref{eq:3} denotes the query feature embedding of the \emph{i}-th primitive, where \emph{n} is the number of points that make up the primitive (e.g., a line consists of two points and a corner consists of three points).
\begin{equation}
   query_{i}=\{query^{k}_{i}|k=1,...,n\}
  \label{eq:3}
\end{equation}

Each group of queries shares a common query position embedding, which is then repeated \emph{n} times to match the number of points in the primitive before the queries are inputted to the decoder block:
\begin{equation}
\label{eq:4}
\left\{
\begin{aligned}
   &Q_{0}=Flatten(\{query_{i}|i=1,...,N\}) \\
   &Q_{0}^{pos}=Flatten(\{Repeat(query_{i}^{pos})|i=1,...,N\})
\end{aligned}
\right.,
\end{equation}
where $query^{pos}_{i} \in \mathbb{R}^{C}$ is the query position embedding for the \emph{i}-th primitive, $N$ is the preset number of primitives making up a building contour. The query input $Q_{0} \in \mathbb{R}^{Nn\times C}$ and the query position input $Q_{0}^{pos} \in \mathbb{R}^{Nn\times C}$ are used as inputs to the decoder block.

\begin{figure*}[t]
\centering
\includegraphics[width=0.75\linewidth]{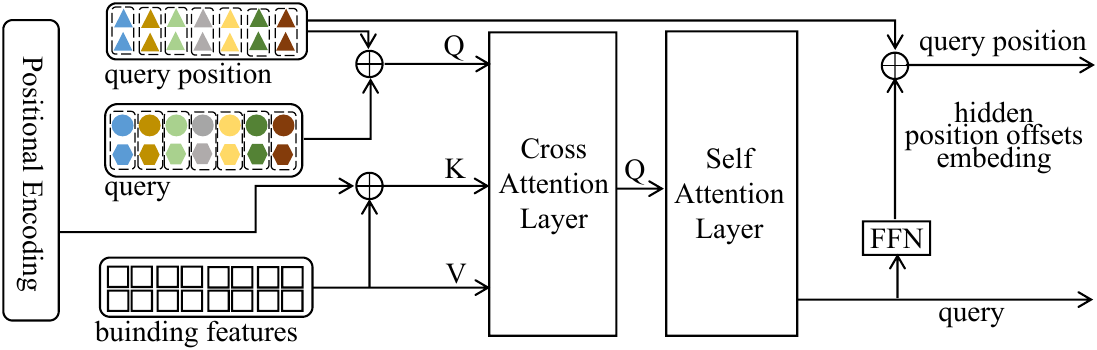}
\caption{The architecture of the primitive decoder block.}
\label{fig:decoder block}
\end{figure*}

\noindent\textbf{Primitive Decoder Block.} The structure of the primitive decoder block, composed of standard cross-attention and self-attention, is depicted in Figure \ref{fig:decoder block}. In this block, the query feature interacts with the instance features using cross-attention, while interaction between query features is achieved using self-attention. At the end, the query position embedding is updated by predicting the implicit offset vector through the FFN:
\begin{equation}
\label{eq:6}
\left\{
\begin{aligned}
   &Q_{l+1}=SA(CA(Q_{l},Q_{l}^{pos},K_{l},V_{l})) \\
   &Q_{l+1}^{pos}=FFN(Q_{l+1})+Q_{l}^{pos} \\
   &K_{l} = f^{decoder}_l, V_{l} = f^{decoder}_l \\
\end{aligned}
\right.,
\end{equation}
\textcolor{black}{where $CA$ is cross-attention, $SA$ is self-attention, l ($l\ge 1$) is the block index, and $f^{decoder}_l$ is the building feature generated from Equation \ref{eq:1}.}

\begin{figure}[t]
\centering
\includegraphics[width=0.65\linewidth]{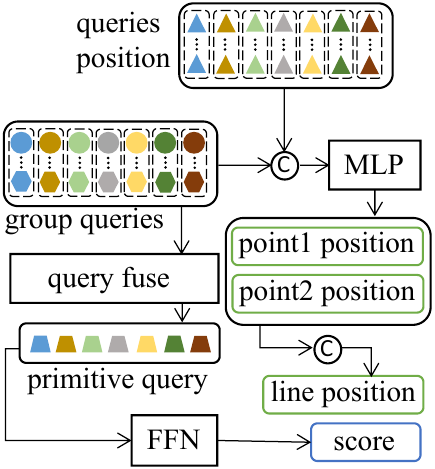}
\caption{The structure of the primitive predictor (using line primitive as the example).}
\label{fig:primitive predictor}
\end{figure}

\noindent\textbf{Primitive Predictor.} The structure of the primitive predictor, exemplified by the line primitive, is illustrated in Figure \ref{fig:primitive predictor}. Utilizing the queries and corresponding position embeddings, the Multilayer Perceptron (MLP) predicts the positions of the two endpoints. The position of the line is subsequently determined by directly combining the positions of the points associated with the current primitive. The Feed-Forward Network (FFN) generates confidence scores based on the fusion of the queries related to the current primitive.
\begin{equation}
\label{eq:9}
\left\{
\begin{aligned}
   &P^{prim}=Reshape(MLP(Concate(Q,Q^{pos}))) \\
   &Q^{prim}=FFN(Reshape(Q)) \\
   &S^{prim}=FFN(Q^{prim})
\end{aligned}
\right.,
\end{equation}
where $Q\in \mathbb{R}^{Nn\times C}$ and $Q^{pos}\in \mathbb{R}^{Nn\times C}$ are the output query and output query position embedding of the last primitive decoder block, respectively. The predicted position of the primitive is represented by $P^{prim}\in \mathbb{R}^{N\times 2n}$, where $n$ is the points number of the primitive. The fused primitive query is denoted as $Q^{prim}\in \mathbb{R}^{N\times C}$, and the predicted confidence score of the primitive is represented by $S^{prim}\in \mathbb{R}^{N\times 2}$ .

\subsection{Order Decoder}

\begin{figure*}[t]
\centering
\includegraphics[width=0.70\linewidth]{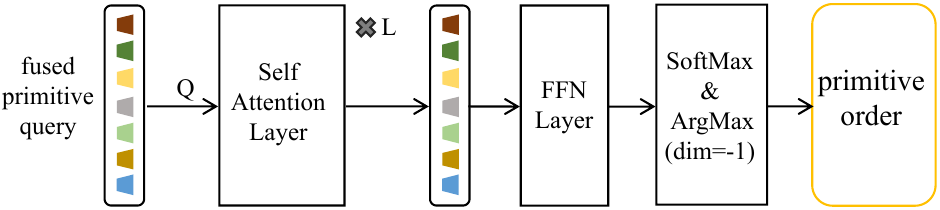}
\caption{The structure of the order decoder.}
\label{fig:order decoder}
\end{figure*}

The architecture of the order decoder is illustrated in Figure \ref{fig:order decoder}. The order decoder effectively utilizes three self-attention layers (L = 3) to model the relationships between primitives. The final order of the primitives is generated by classifying them into predefined order categories based on their representations. The detailed process will be introduced in the following sections.

\noindent\textbf{Primitive Order Label Generation.} We treat primitive order prediction as a classification problem. Specifically, a fixed number of order categories $N_{order}$ are preset, which usually exceeds the number of preset primitives. In this paper, we set $N_{order}$ to 36. The key issue then becomes how to generate the corresponding order category for each ground-truth primitive. To address this, we first discretize the ground-truth building contour into $N_{order}$ points based on the sampling algorithm in \cite{e2ec}. Then, we assign the order of each ground-truth point (clockwise, starting at the 12 o’clock direction) to the order category of the nearest primitive. When using a line as a primitive, the distance is measured between the center point of the line and the sampled ground-truth point. When using a corner as a primitive, the distance is measured between the midpoint of the corner and the sampled point. It should be mentioned that in this paper a corner is defined with three points, the midpoint is the current corner point, and the two endpoints are the two adjacent corner points.

\noindent\textbf{Order Decoder.} The order decoder is composed of a series of self-attention layers. The input query for the order decoder is the $Q^{prim}$, which is outputted by the primitive decoder. The order classification results are derived from the query output by the Feed-Forward Network (FFN). The index corresponding to the maximum response value is utilized as the order index prediction for the primitive.

\section{Implementation Details}
\label{sec:implementation}

\subsection{Detector.}

P2PFormer can use any detector to generate the bboxes of the buildings. In this study, FCOS \cite{fcos} is used as the detector. The P3, P4, and P5 features are used for building detection, and the corresponding regression range is set to (0, 128),(128, 256), and (256, INF).

\subsection{Losses.}

The loss function of P2PFormer consists of a detection loss, a primitive segmentation loss, and a primitive order regression loss:
\begin{equation}
   \mathcal{L}=\mathcal{L}_{det}+\alpha\mathcal{L}_{seg}+\beta\mathcal{L}_{order}
  \label{eq:17}
\end{equation}
where $\mathcal{L}_{det}$ is the detection loss used in \cite{fcos}, $\mathcal{L}_{seg}$ is the primitive segmentation loss, and $\mathcal{L}_{order}$ is the order regression loss. In this study, $\alpha$ was set as 1.0 and $\beta$ was set to 0.1.

\noindent\textbf{Primitive Segmentation Loss.} For each building, $N$ primitives will be segmented. In this study, $N$ is set to 30. $N$ is generally set as greater than the number of ground-truth targets, so that a bipartite matching of the predictions and the ground-truth targets needs to be undertaken to assign a label to each prediction:
\begin{equation}
\label{eq:18}
\left\{
\begin{aligned}
   \mathcal{C} = &sum^{N}_{i=1} \mathds{1}( \sigma(i) \le M) \times \\ &(\lambda_1 L_{1}(P^{prim}_{i}, T^{prim}_{\sigma(i)}) - \lambda_2  S_{i}^{prim}) \\
   \sigma^{*}=&\arg \underset{\sigma}{\min} \, \mathcal{C} \\
\end{aligned}
\right.,
\end{equation}
\textcolor{black}{where $P^{prim}$ is the position prediction of the primitives, $T^{prim}$ is the ground-truth targets, $S^{prim}$ is the score prediction of the primitives, $M$ is the number of ground-truth targets, $\mathds{1}(.)$ is an indicator function, $\sigma(.):\mathbb{Z}_{+}\rightarrow \mathbb{Z}_{+}$ is a map function, and $L_{1}(.,.)$ is the $L_{1}$ distance function. $\lambda_1$ is set to 5.0 and $\lambda_2$ is set to 1.0, which are the same as in \cite{letr}.}

$\mathcal{L}_{seg}$ consists of the classification loss and the regression loss:
\begin{equation}
\label{eq:20}
\left\{
\begin{aligned}
   &\mathcal{L}_{seg}=\lambda_{1}\mathcal{L}_{reg}+\lambda_{2}\mathcal{L}_{cls} \\
   &\mathcal{L}_{cls}=\frac{1}{N}\sum^{N}_{i=1}CE(S_{i}^{prim},\mathds{1}(\sigma^{*}(i)\le M)) \\
   &\mathcal{L}_{reg}=\frac{1}{N}\sum^{N}_{i=1}L_{1}(P_{i}^{prim},T^{prim}_{\sigma^{*}(i)})\mathds{1}(\sigma^{*}(i)\le M) \\
\end{aligned}
\right.,
\end{equation}
where $CE(.,.)$ is the cross entropy loss, and $L_{1}$ is the $L_{1}(.)$ distance loss.

\noindent\textbf{Primitive Order Regression Loss.} We use a simple Cross-entropy loss to supervise the consistency between the predicted and label order categories:
\begin{equation}
\label{eq:23}
\mathcal{L}_{order} = \frac{1}{N}\sum^{N}_{i=1}(CE(O^{prim}_{i}, O^{gt}_{i}) \\,
\end{equation}
where $O^{prim}$ is the predicted order of the primitive, and $O^{gt}$ is the corresponding ground-truth order.

\section{Experiments}
\label{sec:experiments}

\begin{table}[t]
 \centering
 \setlength{\tabcolsep}{0.9mm}
 \caption{\textcolor{black}{Results on the CrowdAI dataset. ``Vector" indicates whether the network can output vector building contours. ``P.F." denotes whether the method is post-processing free. The post-processing includes point NMS, the Douglas–Peucker algorithm, and mask-to-vector transforming algorithms, among others. Note that PolyWorld requires a search algorithm to determine the order of vertices from a large vertex connection matrix.}}
 \begin{tabular}{l|cc|c|ccc}
    \toprule[0.2em]
	Method & Vector & P.F. & Backbone & $mAP$ & $AP_{50}$ & $AP_{75}$\\ \hline
	Mask R-CNN\cite{maskrcnn} & \small{\XSolidBrush} & \small{\XSolidBrush} & ResNet-50 & 41.9 & 67.5 & 48.8\\
	PANet\cite{panet} & \small{\XSolidBrush} & \small{\XSolidBrush} & ResNet-50 & 50.7 & 73.9 & 62.6\\
	Mask2Former\cite{mask2former} & \small{\XSolidBrush} & \small{\XSolidBrush} & ResNet-50 & 63.0 & 91.5 & 72.9 \\
        HigherNet-DST~\cite{he2024highernet} &  \small{\XSolidBrush} & \small{\XSolidBrush} & HigherHRNet & 68.5 & 88.4 & 77.5 \\
	\hline
        PolyMapper\cite{polymapper}  & \small{\CheckmarkBold} & \small{\CheckmarkBold} & ResNet-50 & 55.7 & 86.0 & 65.1\\
	BuildMapper\cite{buildmapper} &  \small{\CheckmarkBold} & \small{\XSolidBrush}  & DLA-34 & 63.9 & 90.1 & 75.0\\
        FFL\cite{ffl} & \small{\CheckmarkBold} & \small{\XSolidBrush} & UR101 & 61.7 & 87.6 & 71.4\\
	PolyWorld\cite{polyworld} & \small{\CheckmarkBold}  & \small{\XSolidBrush}  & R2UNet & 63.3 & 88.6 & 70.5\\
    HiT\cite{zhang2024hit} & \small{\CheckmarkBold} & \small{\CheckmarkBold} & ResNet-50 & 64.6 & 91.9 & 78.7 \\
    
     \hline
	P2PFormer& \small{\CheckmarkBold} & \small{\CheckmarkBold} & ResNet-50~\cite{resnet} & 66.0 & 91.1 & 77.0\\
	P2PFormer&  \small{\CheckmarkBold} & \small{\CheckmarkBold} & Swin-L~\cite{swin} & \textbf{78.3} & \textbf{94.6} & \textbf{87.3} \\
 \bottomrule[0.1em]
 \end{tabular}
 \label{tab:crowdai}
\end{table}

\begin{table}[t]
 \centering
 \setlength{\tabcolsep}{1.6mm}
 \caption{\textcolor{black}{Results on the WHU dataset. ``Vector" indicates whether the network can output vector building contours. ``P.F." denotes whether the method is post-processing free.}}
 \begin{tabular}{l|cc|c|ccc}
    \toprule[0.2em]
     \multirow{2}{*}{Method} & \multirow{2}{*}{Vector} & \multirow{2}{*}{P.F.} & \multirow{2}{*}{Backbone} & \multicolumn{3}{c}{Test}\\
	~ & ~ & ~ & ~ & $mAP$ & $AP_{50}$ & $AP_{75}$ \\
	\hline
	Mask-RCNN\cite{maskrcnn} &  \small{\XSolidBrush} & \small{\XSolidBrush} & ResNet-50 &63.3& 82.9 & 73.9\\
	QueryInst\cite{queryinst} &  \small{\XSolidBrush} & \small{\XSolidBrush} & ResNet-50 &62.6& 78.6 & 71.1\\
	YOLACT\cite{yolact} &  \small{\XSolidBrush} & \small{\XSolidBrush} & ResNet-50 & 58.5 & 76.5 & 69.8\\
	SOLO\cite{solo} &  \small{\XSolidBrush} & \small{\XSolidBrush} & ResNet-50 &  67.9 & 87.5 & 79.2\\
	\hline
	E2EC\cite{e2ec} &  \small{\CheckmarkBold} & \small{\XSolidBrush} & ResNet-50 & 71.3 & 90.4 & 81.4\\
     \hline
	P2PFormer&  \small{\CheckmarkBold} & \small{\CheckmarkBold} & ResNet-50  & \textbf{72.7} & \textbf{91.1} & \textbf{83.2} \\
    \bottomrule[0.1em]
\end{tabular}
\label{tab:whu}
\end{table}
\subsection{Datasets}

We conduct experiments with P2PFormer on the CrowdAI \cite{crowdai}, WHU \cite{mafcn} and WHU-Mix (vector) \cite{buildmapper} datasets.

CrowdAI is a challenging satellite dataset for building segmentation, which contains 280,741 training images and 60,317 test images, with the size of all the images being 300$\times$300 pixels.

The WHU dataset is made up of high-quality aerial images and annotations. The images numbers for the training, validation and test sets are 2793, 627 and 2220, respectively, with the size of each image being 1024$\times$1024 pixels.

The WHU-Mix (vector) dataset (short as WHU-Mix below) is an large aerial and satellite mixed dataset with COCO \cite{coco} format building boundary annotations, and images from more than 10 regions over the world. The WHU-Mix dataset provides two test sets (test1 and test2), with the images of test1 being from similar regions to the training set, and the images of test2 being from completely different cities. The training set has 43,778 images, the validation set has 2922 images, the test1 set has 11,675 images, and test2 set has 6011 images.

\begin{table*}[t]
 \centering
\setlength{\tabcolsep}{3.2mm}
\caption{\textcolor{black}{Results obtained on the WHU-Mix dataset. ``Vector" indicates whether the network can output vector building contours. ``P.F." denotes whether the method is post-processing free.}}
\begin{tabular}{l|cc|c|ccc|ccc}
    \toprule[0.2em]
     \multirow{2}{*}{Method}& \multirow{2}{*}{Vector} & \multirow{2}{*}{P.F.} & \multirow{2}{*}{Backbone} & \multicolumn{3}{c|}{Test1} & \multicolumn{3}{c}{Test2}\\
	~ &  ~ & ~ & ~ & $mAP$ & $AP_{50}$ & $AP_{75}$ & $mAP$ & $AP_{50}$ & $AP_{75}$ \\
	\hline
	Mask R-CNN\cite{maskrcnn} &  \small{\XSolidBrush} & \small{\XSolidBrush} & ResNet50 & 47.0 & 67.0 & 53.2 & 46.1 & 73.9 & 49.0\\
	YOLACT\cite{yolact} &  \small{\XSolidBrush} & \small{\XSolidBrush} & ResNet50 & 42.3 & 65.7 & 47.2 & 41.3 & 71.3 & 42.3\\
	SOLO\cite{solo} &\small{\XSolidBrush} & \small{\XSolidBrush} & ResNet50 & 57.1 & 83.2 & 65.1 & 45.3 & 74.3 & 47.9\\
	Mask2Former\cite{mask2former} & \small{\XSolidBrush} & \small{\XSolidBrush} & ResNet50 & 57.6 & 84.9 & 64.6 & 48.2 & 78.1 & 51.1\\
	\hline
	PolarMask\cite{polarmask} &  \small{\CheckmarkBold} & \small{\XSolidBrush} & ResNet50 & 44.8 & 69.1 & 50.7 & 39.1 & 66.3 & 40.5\\
	Deep Snake\cite{deepsnake} &  \small{\CheckmarkBold} & \small{\XSolidBrush} & DLA-34 & 55.3 & 82.1 & 63.0 & 46.9 & 73.9 & 51.5\\
	BuildMapper\cite{buildmapper} &  \small{\CheckmarkBold} & \small{\XSolidBrush} & ResNet50 & 58.0 & 82.6 & 65.9 & 48.1 & 73.2 & 52.0\\
        Line2Poly\cite{wei2024lines} &  \small{\CheckmarkBold} & \small{\XSolidBrush} & DLA-60 & 58.6 & 81.9 & 66.2 & 48.9 & 73.3 & 52.8 \\
     \hline
	P2PFormer&  \small{\CheckmarkBold} & \small{\CheckmarkBold} & ResNet50 & 60.6 & 87.3 & 68.9 & 50.7 & 79.9 & 54.4 \\
	P2PFormer&  \small{\CheckmarkBold} & \small{\CheckmarkBold} & DLA-34~\cite{dla} & \textbf{61.3} & \textbf{87.6} & \textbf{69.8} & \textbf{53.2} & \textbf{80.9} & \textbf{58.0} \\
    \bottomrule[0.1em]
 \end{tabular}
 \label{tab:whu-mix}
\end{table*}
\subsection{Main Results}
\noindent\textbf{Experimental Settings.} Our primary experiments utilize corners as primitives. P2PFormer employs a standard backbone, pre-trained on ImageNet, to extract image features. It then uses three layers of deformable attention \cite{deformabledetr} to integrate multi-scale image features. Detection is carried out on the multi-scale features from P3-P5, while primitive segmentation is conducted on the P2 feature. For each building, we set the number of queries to 30 and the size of the instance features, cropped by ROI-Align, to 32$\times$32. Given that P2PFormer segments the primitives within the ROI, it is necessary to appropriately expand the ROI to mitigate the adverse effects of detection errors. We set the expansion ratio of the ROI size to 1.1 and apply augmentation during training, randomly chosen from [1.0, (expansion ratio - 1.0) $\times$ 2 + 1.0]. We configure the number of primitive decoder blocks to 3. We train P2PFormer end-to-end for all datasets using the AdamW optimizer with an initial learning rate of 1e-4. For the WHU dataset, we train the model for 50 epochs with a batch size of 4, reducing the learning rate by 0.1 at the 45th epoch. For the CrowdAI and WHU-MIX datasets, we train the model for 24 epochs with a batch size of 16, decreasing the learning rate by 0.1 at the 18th epoch.

\noindent\textbf{Performance on the CrowdAI Dataset.} Table \ref{tab:crowdai} shows the results of various building segmentation methods on the CrowdAI dataset. When using ResNet50~\cite{resnet} as the backbone, P2PFormer achieves a new SOTA performance, outperforming the current SOTA method PolyWorld by 2.7 AP and 6.5 AP$_{75}$. Furthermore, P2PFormer also surpasses BuildMapper (an improved version of E2EC \cite{e2ec} for building segmentation) by 2.1 AP. In addition, P2PFormer achieves 4.1 AP higher than FFL with a simpler and more elegant pipeline.

\noindent\textbf{Performance on the WHU Dataset.} Table \ref{tab:whu} presents the performance comparison of different methods on the WHU dataset. It is evident that P2PFormer outperforms the current SOTA mask-based method by 4.8 AP and the SOTA contour-based method by 1.4 AP. Moreover, P2PFormer can directly predict the regular building contours, which is not feasible for both mask-based and contour-based methods. In this study, we use FCOS as the detector, the same as E2EC, and achieve essentially the same detection performance as E2EC on the WHU test set (75.1 $vs.$ 75.0). Therefore, the performance gain of 1.4 AP entirely comes from the proposed primitive-based segmentation method.

\noindent\textbf{Performance on the WHU-Mix Dataset.} Table \ref{tab:whu-mix} shows the performance of different methods on the WHU-Mix dataset. P2PFormer outperforms the current SOTA method BuildMapper by 2.6 AP and 4.7 AP$_{50}$ on the test1 set, and 2.6 AP and 6.7 AP$_{50}$ on the test2 set when using ResNet50. The excellent performance of P2PFormer on the WHU-Mix dataset indicates that the proposed method can handle diverse image styles, instance styles, and source regions while still achieving high accuracy.

\begin{table}[t]
\centering
\setlength{\tabcolsep}{4.0mm}
\caption{Performance of P2PFormer with different primitive types. The ablation study is conducted on the CrowdAI dataset using the Swin-L backbone. For all primitive types, the network and training settings are kept consistent.}
\begin{tabular}{l|cccc}
    \toprule[0.2em]
	Primitive Type & $mAP$ & $AP_{50}$ & $AP_{75}$  & $AR$\\
	\hline
	vertex primitive & 72.6 & 87.6 & 79.9  & 76.4 \\
	line primitive & 75.0 & 90.0 & 83.2 & 79.3 \\
	corner primitive & \textbf{78.3} & \textbf{94.6} & \textbf{87.3}&  \textbf{81.8}\\
\bottomrule[0.1em]
\end{tabular}
 \label{tab:primitive types}
\end{table}

\begin{table}[t]
 \centering
 \setlength{\tabcolsep}{1.2mm}
 \caption{Results of different primitive representation strategies. The ablation study is conducted on the WHU dataset using a ResNet-50 backbone. The P2PFormer employs lines as primitives, with all other settings identical to those in the main experiments.}
 \begin{tabular}{cc|ccc|ccc}
    \toprule[0.2em]
	\multirow{2}{*}{Group Que.} & \multirow{2}{*}{Group Pos.} &  \multicolumn{3}{c|}{Val} & \multicolumn{3}{c}{Test}\\
	~ & ~ & $mAP$ & $AP_{50}$ & $AP_{75}$ & $mAP$ & $AP_{50}$ & $AP_{75}$ \\
	\hline
	\multicolumn{2}{l|}{DETR mode:} &&&&&\\ 
      \small{\XSolidBrush} & \small{\XSolidBrush} & 72.9 & 91.0 & 81.7 & 70.4 & 88.8 & 79.9\\
	\hline
	\multicolumn{2}{l|}{Our proposed:}&&&&&\\ 
	\small{\XSolidBrush} & \small{\CheckmarkBold} & 74.2 & 91.5 & 83.4 & 71.5 & 90.1 & 81.4\\
	\small{\CheckmarkBold} & \small{\XSolidBrush} & \textbf{75.1} & \textbf{91.9} & \textbf{84.7} & 72.5 & 90.2 & \textbf{82.7}\\
	\small{\CheckmarkBold} & \small{\CheckmarkBold} & 75.0 & 91.7 & 84.7 & \textbf{72.6} & \textbf{90.3} & 82.2\\
 \bottomrule[0.1em]
 \end{tabular}
 \label{tab:query mode}
\end{table}

\subsection{Ablation Experiments}

\noindent\textbf{Primitive Type.} In order to investigate the impact of primitive types on the upper performance limit of our model, we utilize vertex, line, and corner as primitives, with each primitive consisting of 1, 2, and 3 points, respectively, as a simplest building has at least 3 points. The performance of P2PFormer, with Swin-L as the backbone, using different primitive types on the largest CrowdAI dataset is presented in Table \ref{tab:primitive types}. Our results show that corner primitives achieve the best performance, with significantly higher AP and AR compared to vertex primitives. This is because the probability of occlusion of corner primitives is much lower than that of vertex primitives, which results in higher recall and more accurate regular building contours. Missing primitives can be fatal for primitive-based methods, thus the use of corner primitives leads to more robust and accurate segmentation results.

\noindent\textbf{Group Queries for Primitives Representation.} The standard DETR-style representation performs well for vertices, but has non-negligible shortcomings for lines or corners. For instance, \cite{letr} employed the standard DETR-style representation with a single query for each line, but the prediction of line endpoint positions is often inaccurate due to a single query's difficulty in focusing on both endpoints simultaneously.

In this paper, we propose representing a primitive using a group of queries $\{Q_{i}|i\in[1,n]\}$, where the queries in the same group share a query position embedding, and $n$ is the number of implicit points (i.e., the parameter number in 2D) in the primitive. The $i$-th point coordinates of the primitive are predicted by $Q_{i}$ individually, and the confidence score of the primitive is predicted by fusing a group of queries to generate $Q_{be}$.

Table \ref{tab:query mode} presents the results of our ablation experiments conducted on the group query representation. Our proposed method of representing a primitive by a group of queries $\{Q_{i}|i\in[1,n]\}$, where the queries belonging to the same group share an query position embedding, results in a gain of 2.2 AP compared to the standard DETR-style representation. We also experiment with different variants of the group query representation, such as representing a primitive with a group of query position embeddings and a shared query with the same initial value, which results in a significant performance degradation but still outperforms the standard DETR-style representation. On the other hand, the representation strategy using a group of queries and a group of query position embeddings, although theoretically more powerful, does not provide any significant gain and instead introduced a performance degradation of 0.5 AP$_{75}$ on the WHU test set, compared to the shared initial query position embeddings.

\begin{table}[t]
\centering
\setlength{\tabcolsep}{2.3mm}
\caption{Results from the use of various query position embedding strategies. The ablation study is conducted on the WHU dataset using a ResNet-50 backbone. The P2PFormer employs lines as primitives, maintaining all other settings identical to those in the main experiments. Performance is evaluated on the validation set.}
\begin{tabular}{c|c|ccc}
    \toprule[0.2em]
	Update Pos. & Predict Pos. with Que. & $mAP$ & $AP_{50}$ & $AP_{75}$ \\
	\hline
	\small{\XSolidBrush} & \small{\XSolidBrush} & 73.8 & 89.5 & 82.4 \\
	\small{\XSolidBrush} & \small{\CheckmarkBold} & 74.4 & 91.8 & 84.3 \\
	\small{\CheckmarkBold} & \small{\XSolidBrush} & 74.7 & 91.6 & 84.3 \\
	\small{\CheckmarkBold} & \small{\CheckmarkBold} & \textbf{75.1} & \textbf{91.9} & \textbf{84.7} \\ 
\bottomrule[0.1em]
 \end{tabular}
 \label{tab:pos strategy}
\end{table}

\begin{table}[t]
\centering
\setlength{\tabcolsep}{1.3mm}
\caption{Results of P2PFormer utilizing features of different sizes for the primitive decoder block. Single-scale feature ($S$) refers to the direct use of ROI-Align to crop features of size $S\times S$ from the image features. Multi-scale features ($S1, S2, S3$) involve the use of ROI-Align to crop features of size $S3\times S3$ from the image features, followed by the generation of features of sizes $S2\times S2$ and $S1\times S1$ using two down-sampling modules.}
 \begin{tabular}{l|ccc|ccc}
   \toprule[0.2em]
	\multirow{2}{*}{Features} & \multicolumn{3}{c|}{Val} & \multicolumn{3}{c}{Test} \\
	~ & $mAP$ & $AP_{50}$ & $AP_{75}$ & $mAP$ & $AP_{50}$ & $AP_{75}$ \\
	\hline
	Multi-scale (4,8,16) & 74.4 & 91.7 & 84.6 & 72.1 & 90.3 & 82.6 \\
	Multi-scale (8,16,32) & 75.1 & 91.9 & 84.7 & 72.5 & 90.2 & 82.7 \\
	Multi-scale (16,32,64) & 74.7 & 91.7 & 84.4 & 72.1 & 90.2 & 81.9 \\
	Single-scale (8) & 74.4 & 91.8 & 84.6 & 71.9 & 90.3 & 82.1 \\
	Single-scale (16) & 74.5 & 91.7 & 84.6 & 71.9 & 90.2 & 81.9 \\
	Single-scale (32) & 73.8 & 91.6 & 83.6 & 71.8 & 90.2 & 81.9 \\ 
 \bottomrule[0.1em]
 \end{tabular}
 \label{tab:decoder feature}
\end{table}

\begin{table}[t]
\centering
 \caption{Results of P2PFormer with different numbers of queries. The ablation study is conducted on the WHU dataset using a ResNet-50 backbone. The P2PFormer employs lines as primitives, maintaining all other settings identical to those in the main experiments.}
\setlength{\tabcolsep}{3.0mm}
\begin{tabular}{c|ccc|ccc}
    \toprule[0.2em]
	\multirow{2}{*}{N} & \multicolumn{3}{c|}{Val} & \multicolumn{3}{c}{Test}\\
	~ &  $mAP$ & $AP_{50}$ & $AP_{75}$ & $mAP$ & $AP_{50}$ & $AP_{75}$\\
	\hline
	30 & 75.1 & 91.9 & 84.7 & 72.5 & 90.2 & 82.7 \\
	60 & 75.5 & 91.8 & 84.9 & 72.7 & 90.3 & 82.9 \\
    \bottomrule[0.1em]
 \end{tabular}
 \label{tab:querynum}
\end{table}

\begin{figure*}[t]
\centering
\includegraphics[width=0.98\linewidth]{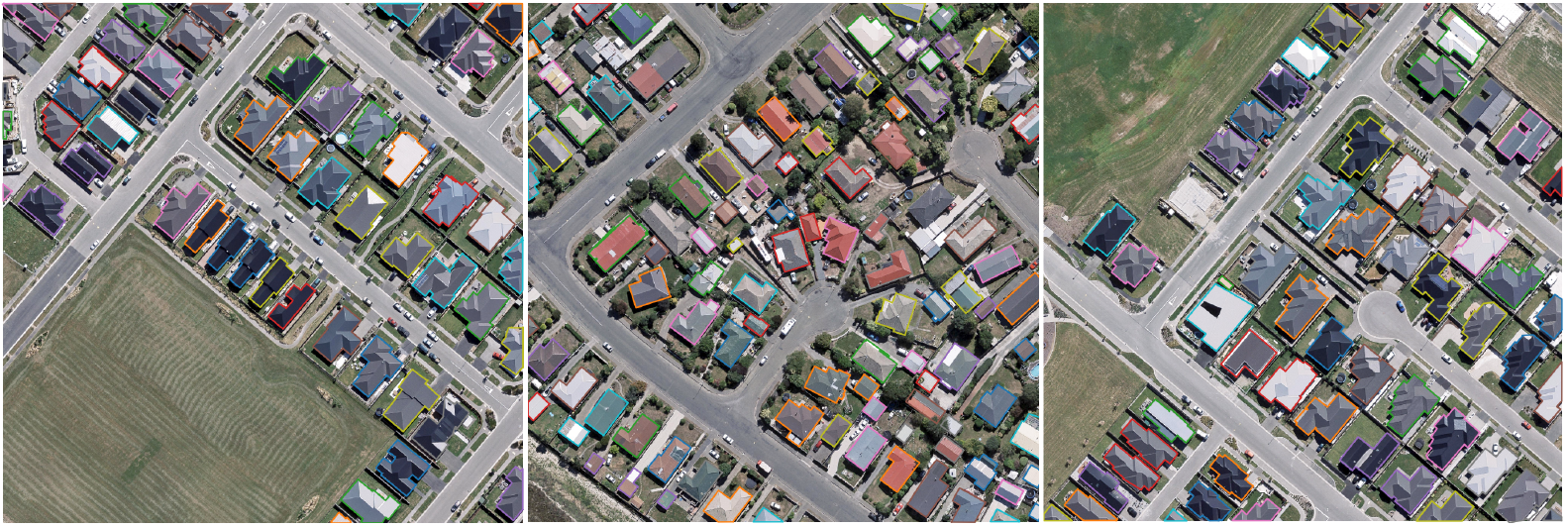}
\includegraphics[width=0.98\linewidth]{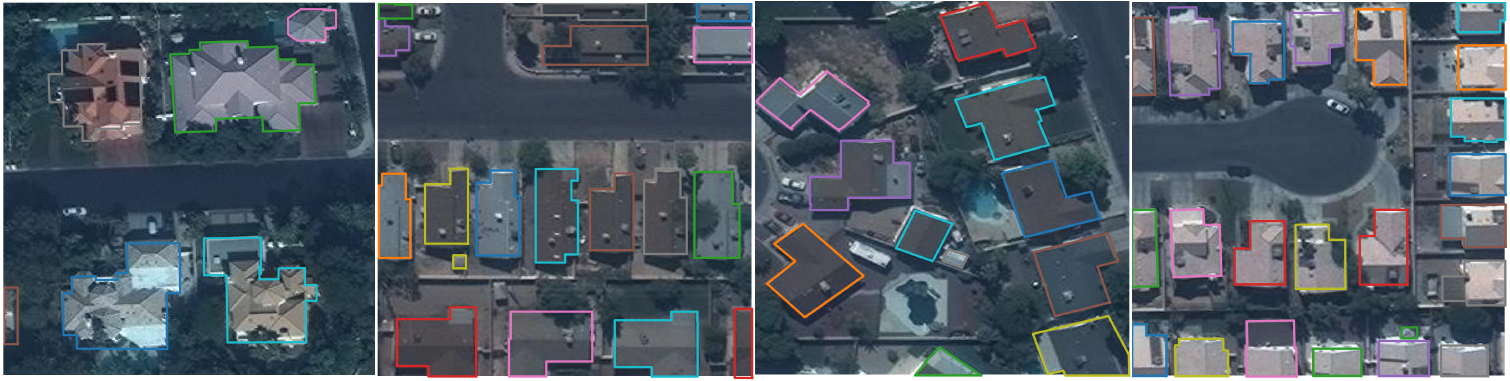}
\includegraphics[width=0.98\linewidth]{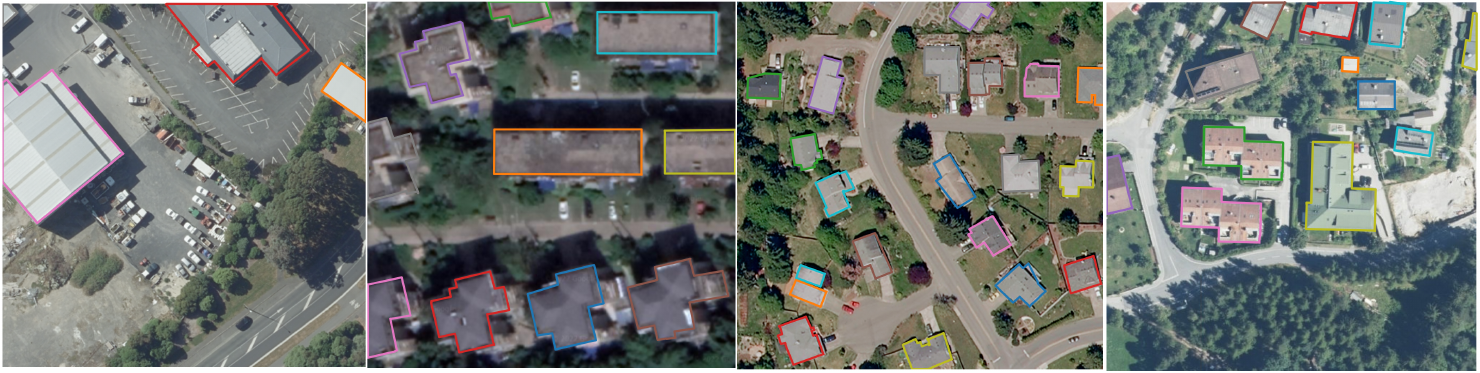}
\caption{Qualitative results obtained by P2PFormer. From top to bottom: results on the WHU, CrowdAI, and WHU-Mix datasets.}
\label{fig:demos}
\end{figure*}

\begin{figure*}[t]
\centering
\includegraphics[width=0.92\linewidth]{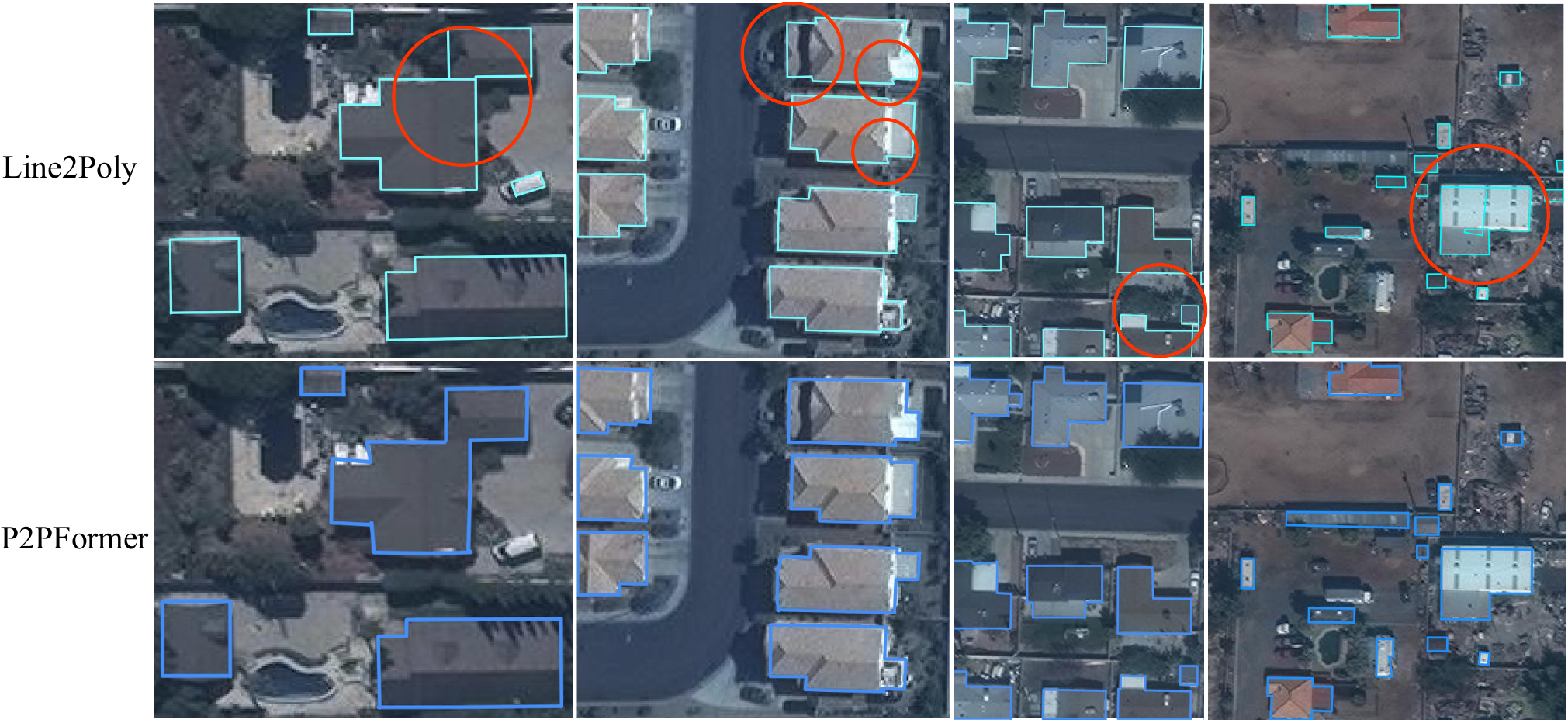}
\includegraphics[width=0.92\linewidth]{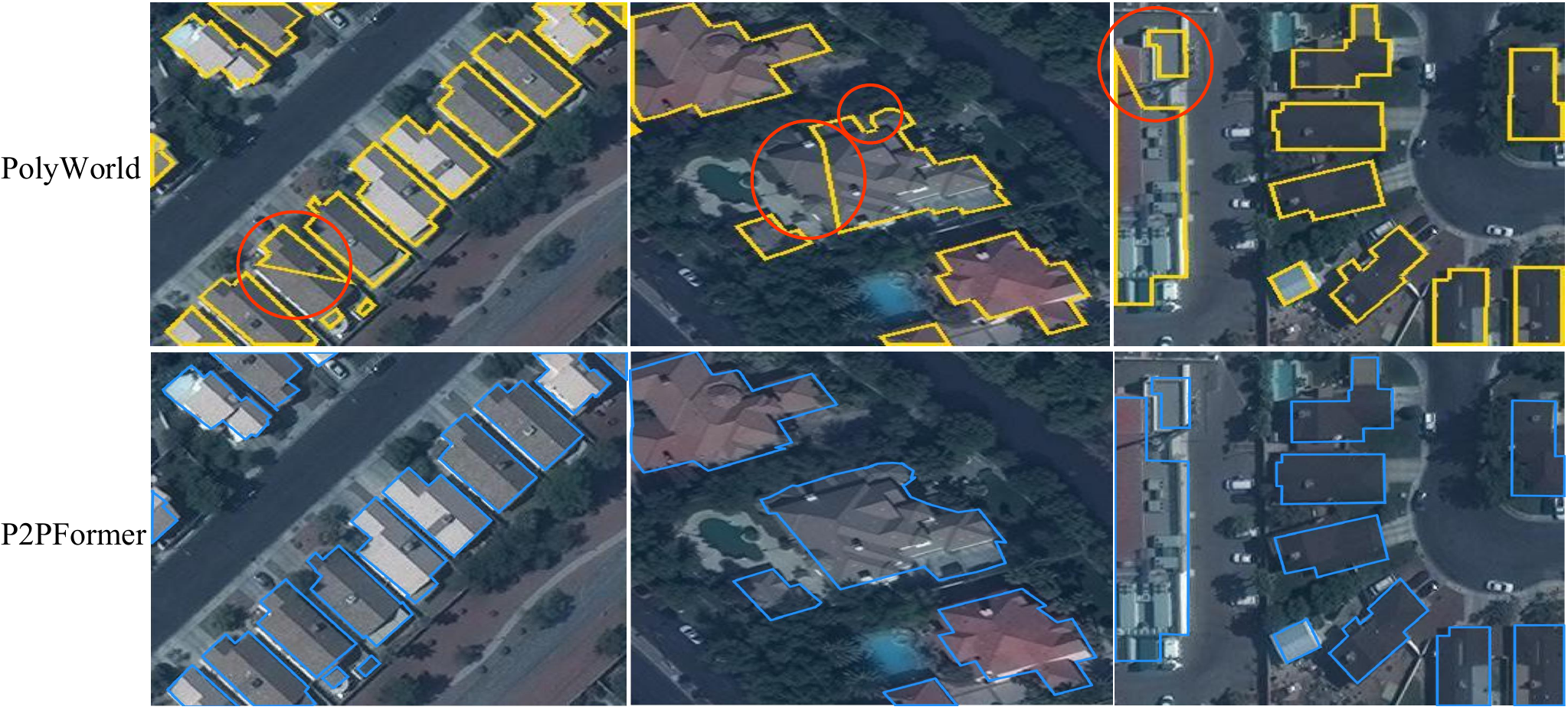}
\caption{\textcolor{black}{Qualitative comparison with other SOTA methods. The deficiencies in the predictions of Line2Poly~\cite{wei2024lines} and PolyWorld~\cite{polyworld} are highlighted with red circles. Compared to Line2Poly, P2PFormer exhibits fewer topological errors. Additionally, compared to PolyWorld, P2PFormer demonstrates more stable primitive segmentation capability and higher contour prediction quality.}}
\label{fig:compare}
\end{figure*}

\noindent\textbf{Query Position Embedding Strategies.} Although the query position embedding contains rich location information, previous works such as \cite{detr,letr} only utilize it to generate positional attention maps. Therefore, we propose a new strategy for updating the position of primitives by fully leveraging the query position embedding. Table \ref{tab:pos strategy} presents the results of ablation experiments conducted for these strategies. Predicting the position of primitives based on both the query and the query position embedding leads to a gain of 0.6 AP. Additionally, applying an implicit dynamic update to the query position embedding helps the model attend to more appropriate locations, resulting in a gain of 0.9 AP. The combined use of these two strategies leads to a total gain of 1.3 AP, 2.4 AP$_{50}$, and 2.3 AP$_{75}$.

\noindent\textbf{Instance Feature Size.} Table \ref{tab:decoder feature} presents the results of the ablation experiments on the instance feature size used in the primitive decoder block. Our experiments show that the use of multi-scale instance features effectively improves the network's performance, with the multi-scale feature (row 2) delivering a gain of at least 0.6 AP compared to using any single scale (rows 4, 5, 6). However, instance features of a smaller size (row 1) suffer from a severe loss of details, resulting in a 0.7 AP drop in performance compared to row 2. On the other hand, features with a larger size (row 3) increase the difficulty of convergence, resulting in a 0.4 AP drop compared to row 2.

\noindent\textbf{Number of Queries.} Table \ref{tab:querynum} lists the experimental results of P2PFormer with different number of queries on the WHU dataset. Doubling the number of queries brings a limited performance gain, so 30 was used as the number of queries in this study, which is sufficient for representing the primitives of various building structures. 

\subsection{Qualitative Analysis}

\begin{figure}[t]
\begin{center}
\begin{minipage}[c]{0.23\linewidth}
\includegraphics[width=0.98\linewidth]{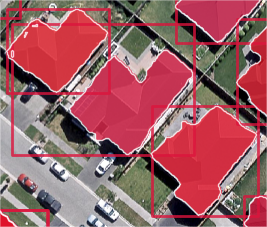}
\end{minipage}\hfill
\begin{minipage}[c]{0.23\linewidth}
\includegraphics[width=0.98\linewidth]{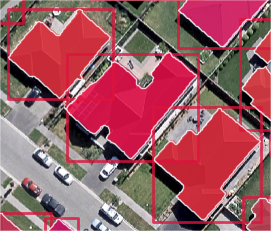}
\end{minipage}\hfill
\begin{minipage}[c]{0.23\linewidth}
\includegraphics[width=0.98\linewidth]{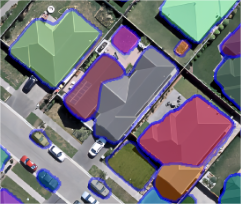}
\end{minipage}\hfill
\begin{minipage}[c]{0.23\linewidth}
\includegraphics[width=0.98\linewidth]{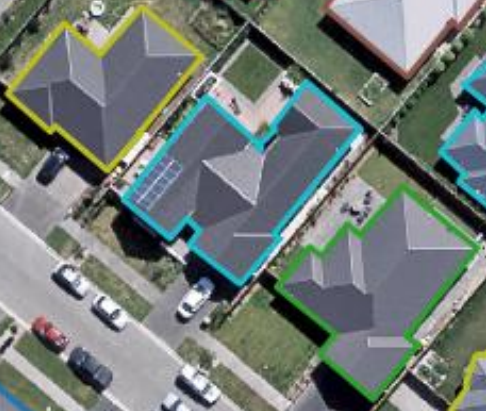}
\end{minipage}\hfill
\caption{Visualization of different methods. From left to right are Mask2Former, E2EC, SAM, and P2PFormer.}
\label{fig:compare}
\end{center}
\end{figure}

\noindent\textbf{Visualization Results.} Figure~\ref{fig:demos} displays the prediction results of P2PFormer on the WHU, CrowdAI, and WHU-Mix datasets. P2PFormer can accurately segment primitives and predict their sequence, demonstrating nearly perfect results in building polygon extraction. The level of regularity is almost equivalent to manual annotation.

\textcolor{black}{Figure~\ref{fig:compare} presents the visualization comparison with other SOTA building extraction methods. Compared to Line2Poly~\cite{wei2024lines}, the predictions of P2PFormer exhibit a more stable topological structure. This improvement is attributed to P2PFormer's novel approach of predicting primitive order instead of the complex and unstable method of predicting vertex connections to determine topological relationships. Compared to PolyWorld~\cite{polyworld}, P2PFormer demonstrates more stable primitive segmentation results and better contour quality.}

\noindent\textbf{Qualitative Comparison with Other SOTA Methods.} Figure \ref{fig:compare} presents the outcomes of applying advanced mask-based and contour-based segmentation methods, such as Mask2Former~\cite{mask2former}, E2EC~\cite{e2ec}, and SAM~\cite{sam}, to building extraction. Nonetheless, their predicted results are inevitably imprecise and tend to be ``rounded". In contrast, only our method attains an accuracy that aligns with human delineation.

\begin{figure}[t]
\begin{center}
\begin{minipage}[c]{0.23\linewidth}
\includegraphics[width=1.00\linewidth]{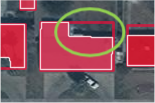}
\end{minipage}\hfill
\begin{minipage}[c]{0.23\linewidth}
\includegraphics[width=1.00\linewidth]{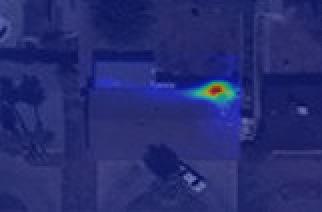}
\end{minipage}\hfill
\begin{minipage}[c]{0.23\linewidth}
\includegraphics[width=1.00\linewidth]{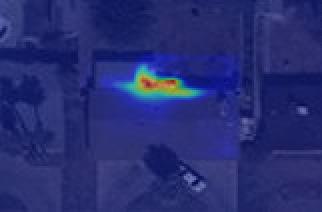}
\end{minipage}\hfill
\begin{minipage}[c]{0.23\linewidth}
\includegraphics[width=1.00\linewidth]{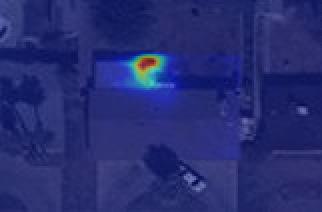}
\end{minipage}\hfill
\caption{Corner primitive prediction and attention maps.}\label{fig:attn}
\end{center}
\end{figure}

\begin{figure}[t]
\begin{center}
\begin{minipage}[c]{0.43\linewidth}
\includegraphics[width=1.00\linewidth]{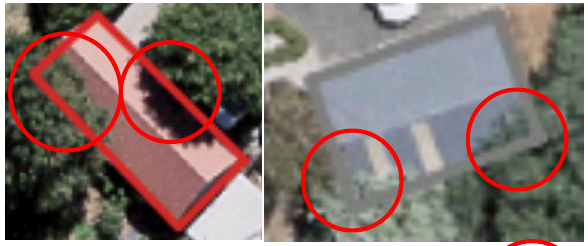}
\end{minipage}\hfill
\begin{minipage}[c]{0.56\linewidth}
\includegraphics[width=1.00\linewidth]{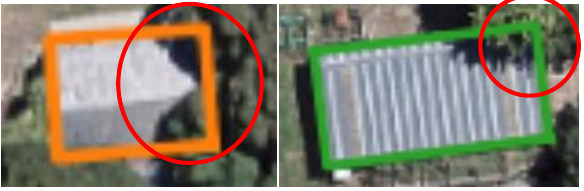}
\end{minipage}\hfill
\caption{Predicted results of occlusion examples. The occluded part of a building is highlighted by a red circle.}\label{fig:occlusion}
\end{center}
\end{figure}

\noindent\textbf{Visualization of Primitive Query Attention Maps.} Figure \ref{fig:attn} presents a predicted result of an corner primitive, along with its group queries and the corresponding attention maps within the image. It can be observed that the three queries representing the corner primitive focus on the three endpoints of the angle. This validates the effectiveness of our proposed group queries representation.

\noindent\textbf{Visualization Results for Occluded Cases.} Figure~\ref{fig:occlusion} demonstrates the extraction outcomes on occluded buildings using the P2PFormer with corner primitives. Corners are less likely to be wholly occluded than points and lines. Therefore, the P2PFormer effectively addresses challenging occlusion scenarios by utilizing corner primitives, which are more difficult to occlude entirely.

\begin{figure}[t]
\begin{center}
\begin{minipage}[c]{0.41\linewidth}
\includegraphics[width=1.00\linewidth]{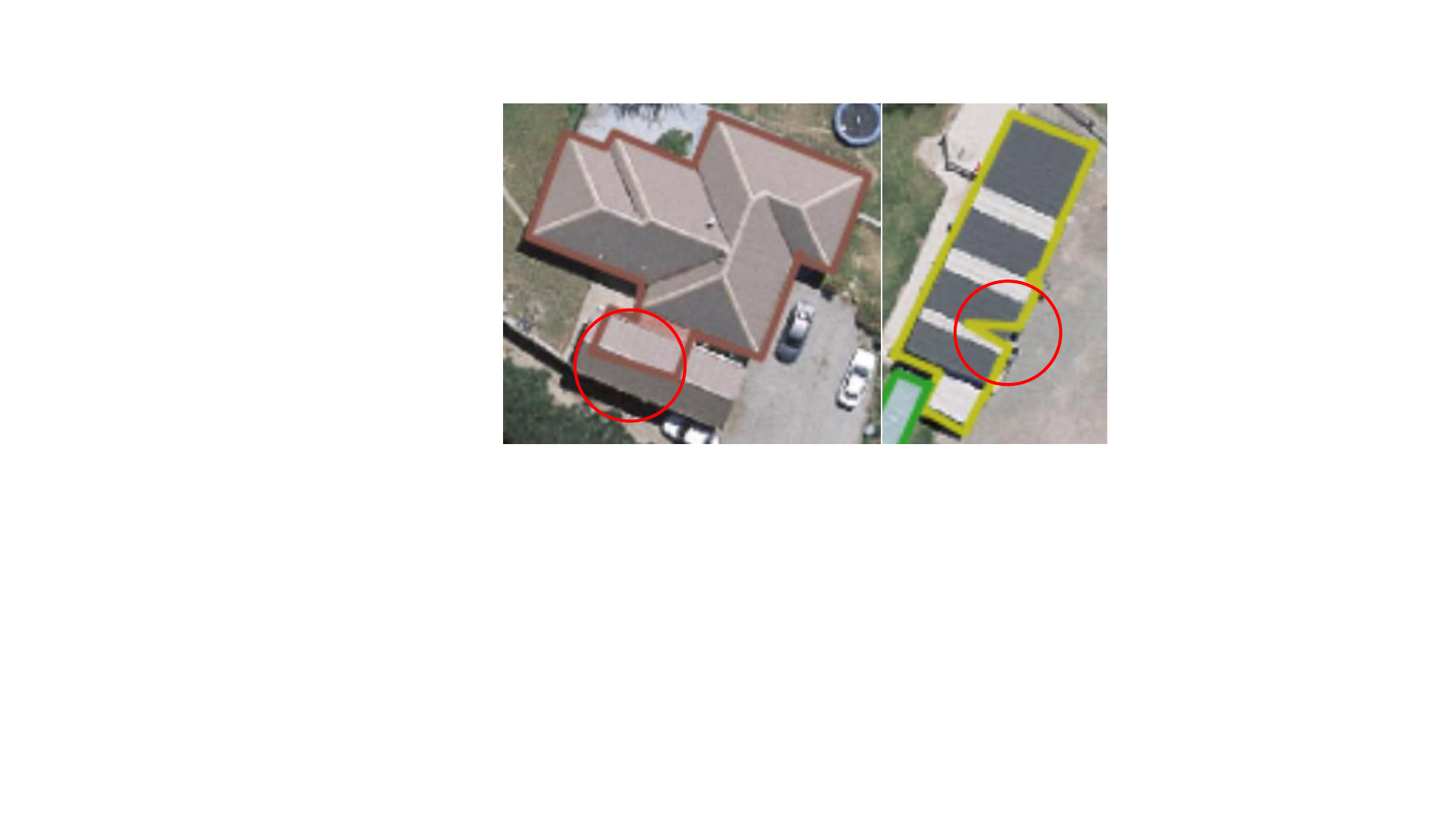}
\end{minipage}\hfill
\begin{minipage}[c]{0.58\linewidth}
\includegraphics[width=1.00\linewidth]{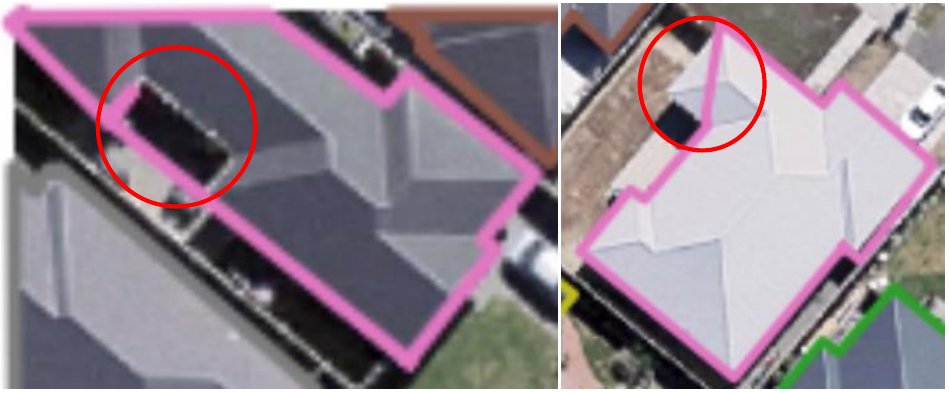}
\end{minipage}\hfill
\caption{Failure cases. The inaccurate part of the building extraction result is highlighted by a red circle.}\label{fig:fail}
\end{center}
\end{figure}

\noindent\textbf{Failure Cases.} Despite P2PFormer adopting a more straightforward approach to achieve state-of-the-art (SOTA) performance in regular building contour extraction, there remains considerable room for improvement. Firstly, the segmentation of primitives can be disrupted by interior primitives of buildings, as illustrated on the left side of Figure~\ref{fig:fail}, where incorrect segmentation of internal primitives leads to the failure of building contour extraction. This challenge might be addressed by masking out the primitive queries' perception on interior building features, which we plan to explore. Secondly, the absence of primitives can also fail in building contour extraction, as shown on the right side of Figure~\ref{fig:fail}. This issue calls for an enhancement of the primitive segmenter's capabilities. Finally, P2PFormer sometimes underperforms on large buildings because building features are extracted from images using ROI-Align with a small fixed grid pattern, which reduces the likelihood of sampling points right on the contours of larger buildings. This challenge necessitates a more flexible approach or a feature extraction method that can correct points towards contours.

\section{Conclusion}
\label{conclusion}
This paper presents a novel and concise pipeline for regular building contour extraction that does not rely on any post-processing. We introduce P2PFormer, which includes a primitive segmenter and a primitive order decoder following the pipeline. Our primitive segmentation module incorporates three innovative designs that significantly improve the quality of primitive segmentation. Leveraging these advantages, P2PFormer achieves new SOTA performance on the CrowdAI, WHU, and WHU-Mix datasets when using corner primitives, enabling the end-to-end extraction of regular building contours. We believe that P2PFormer will serve as a strong and fundamental baseline for regular building contour extraction.



\section{Acknowledgement}

This work is supported by the National Natural Science Foundation of China (grant No. 42171430) and the State Key Program of the National Natural Science Foundation of China (grant No. 42030102).

{
	\bibliographystyle{IEEEtran}
	\bibliography{IEEEabrv,ref}
}


\vfill

\end{document}